\documentclass[10pt,twocolumn,letterpaper]{article}

\usepackage[pagenumbers]{cvpr} % To force page numbers, e.g. for an arXiv version
\usepackage{xcolor}         % colors
\usepackage{float}
\usepackage{placeins} 
\usepackage{cuted}
\usepackage{caption}
\usepackage{subcaption}
\usepackage{graphicx}
\usepackage{multirow}
\usepackage{amsmath}
\usepackage{tablefootnote}
\usepackage[table]{xcolor}
\usepackage{colortbl}
\usepackage{ragged2e}

\definecolor{cvprblue}{rgb}{0.21,0.49,0.74}
\usepackage[pagebackref,breaklinks,colorlinks,allcolors=cvprblue]{hyperref}

\def\paperID{32510} % *** Enter the Paper ID here
\def\confName{CVPR}
\def\confYear{2026}

\title{GridNet-HD: A High-Resolution Multi-Modal Dataset for LiDAR-Image Fusion on Power Line Infrastructure}

\author{
Antoine Carreaud$^{1,2}$\quad
Shanci Li$^{2}$ \quad
Malo De Lacour$^{2}$ \quad
Digre Frinde$^{2}$ \quad\\
Jan Skaloud$^{1}$ \quad
Adrien Gressin$^{2}$\\
$^{1}$ESO lab. EPFL, 1015 Lausanne, Switzerland - (firstname.lastname)@epfl.ch \\
$^{2}$University of Applied Sciences Western Switzerland (HES-SO / HEIG-VD),\\
Yverdon-les-Bains, Switzerland - (firstname.lastname)@heig-vd.ch
}

\begin{document}
\maketitle

\begin{strip}
\vspace{-1.8 cm}
\centering
\captionsetup{type=figure} % makes \caption legal here

\subcaptionbox{LiDAR - RGB.\label{fig:dataset_a}}[0.29\textwidth]{%
  \includegraphics[width=\linewidth]{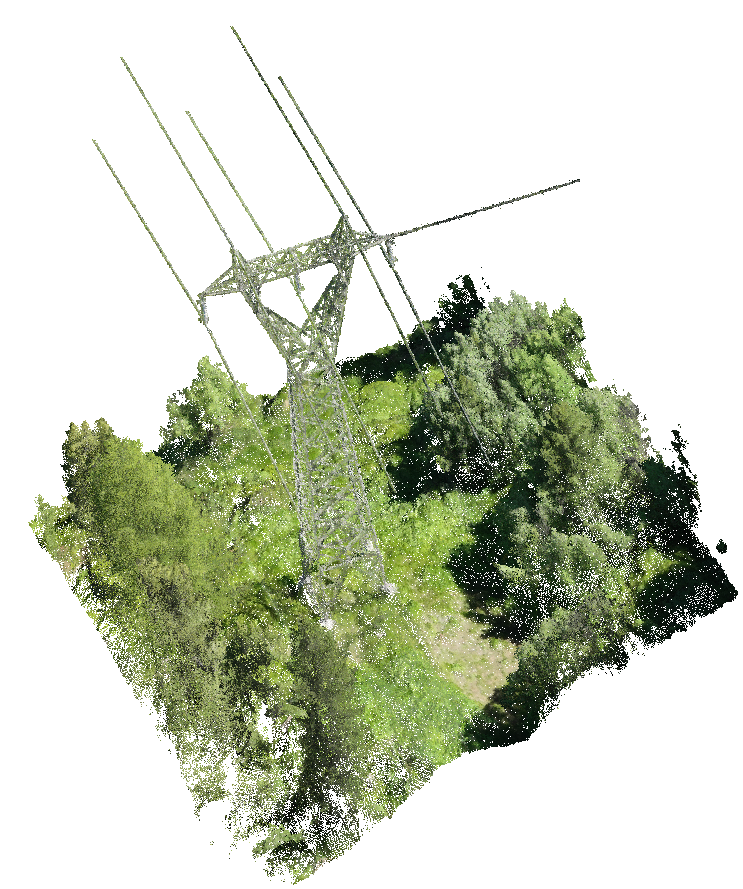}}
\hfill
\subcaptionbox{LiDAR - semantic classes.\label{fig:dataset_b}}[0.29\textwidth]{%
  \includegraphics[width=\linewidth]{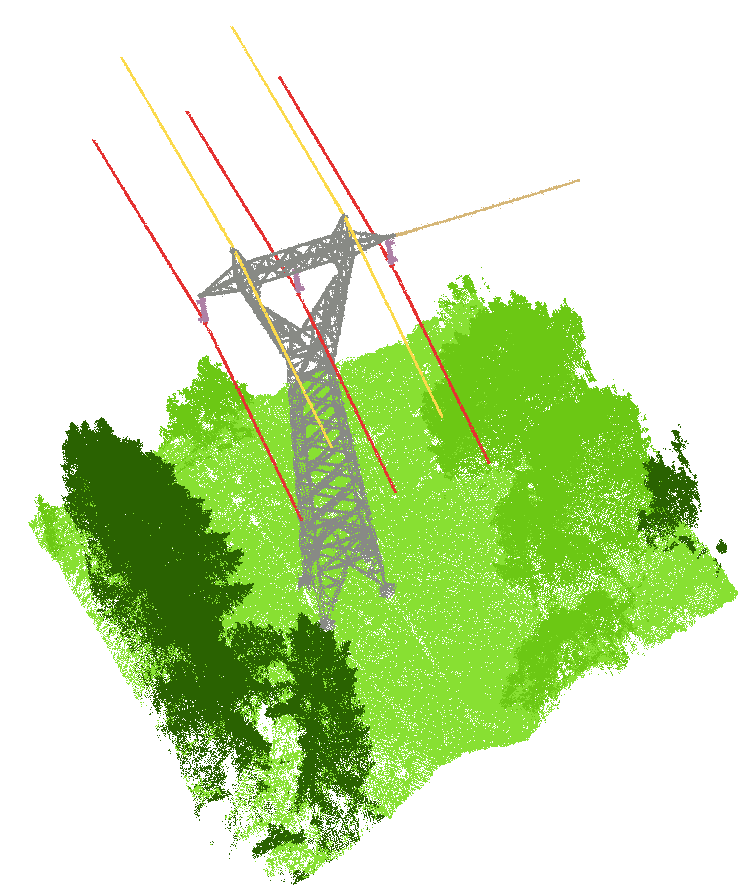}}
\hfill
\subcaptionbox{Image of the area.\label{fig:dataset_c}}[0.29\textwidth]{%
  \includegraphics[width=\linewidth]{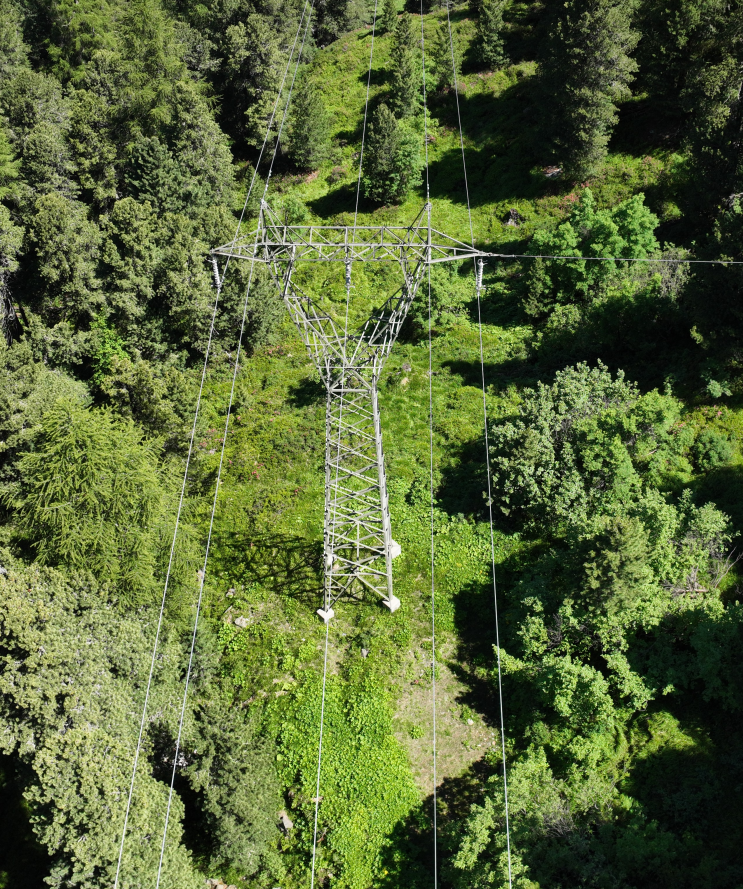}}

\caption{Example of a small area from the dataset. (a) LiDAR data visualized in RGB, (b) LiDAR data colorized by semantic classes, (c) Image of the corresponding area.}
\label{fig:dataset}
\end{strip}

\begin{abstract}
This paper presents GridNet-HD, a multi-modal dataset for 3D semantic segmentation of overhead electrical infrastructures, pairing high-density LiDAR with high-resolution oblique imagery. The dataset comprises 7,694 images and 2.5 billion points annotated into 11 classes, with predefined splits and mIoU metrics. Unimodal (LiDAR-only, image-only) and multi-modal fusion baselines are provided. On GridNet-HD, fusion models outperform the best unimodal baseline by +5.55 mIoU, highlighting the complementarity of geometry and appearance. As reviewed in Sec.~2, no public dataset jointly provides high-density LiDAR and high-resolution oblique imagery with 3D semantic labels for power-line assets. Dataset, baselines, and codes are available: \nolinkurl{https://huggingface.co/collections/heig-vd-geo/gridnet-hd}. 
\end{abstract}

\section{Introduction}
\label{sec:intro}

The fusion of LiDAR and image data has become increasingly prevalent in computer vision, and artificial intelligence, with applications covering a wide range of domains, from autonomous driving and robotics~\cite{DeepFusion, transfuser, BEVFusion, superfusion} to urban mapping~\cite{deepviewagg} and remote sensing~\cite{tuia_lidar}. Thanks to decades of research, LiDAR delivers highly accurate 3D georeferenced spatial information~\cite{GRESSIN_lidar, brun_pointopoint}, essential to capture fine structural details, while images provide rich semantic content~\cite{gressin2020photogrammetric}. Despite these complementary strengths, the accurate alignment of these heterogeneous data sources remains challenging. Photogrammetry alone does not guarantee a perfect one-to-one correspondence between image pixels and LiDAR points. Consequently, extensive research has focused on bridging this gap to achieve increasingly accurate multimodal alignment~\cite{kyriaki, cledat2020, kiriaky_open_journal}.

However, several challenges still persist in multimodal fusion:

\begin{enumerate}
\item \textbf{Data alignment:} Achieving accurate co-registration is complex due to modality differences. This challenge is addressed through a rigorous geospatial alignment workflow, ensuring high-precision registration between images and LiDAR data. The methodology and validation procedures are described in Section~\ref{sec:dataset}.
\item \textbf{Limited availability of suitable datasets:} Despite the growing interest in multimodal fusion, there is a notable lack of large-scale, high-quality datasets tailored for 3D semantic segmentation tasks combining LiDAR and images. Existing benchmarks often focus on a single modality, lack resolution, or do not provide precise alignment. We provide a review of current datasets in Section~\ref{sec:related_work}.
\item \textbf{Fusion mechanisms:} Identifying effective strategies for multimodal fusion remains an open research question. Although our primary contribution is the introduction of a novel dataset, we also establish five baseline methods for 3D semantic segmentation:
\begin{itemize}
    \item An image-only approach that reprojects image-based features into 3D by majority voting,
    \item A SuperPoint Transformer (SPT)~\cite{robert2023spt} and a Point Transformer (PTv3)~\cite{PTv3}, representative of the current state-of-the-art in point cloud segmentation,
    \item Two fusion strategies, a late-fusion combining outputs from image-based and SPT through a Multi-Layer Perceptron (MLP), a feature fusion method based on the current SOTA "Dino In The Room" (DITR)~\cite{DITR}.
\end{itemize}
These approaches serve as benchmarks for evaluating performance and are described in Section~\ref{sec:baseline}.
\end{enumerate}

In the electrical infrastructure sector, high-density LiDAR and high-resolution oblique imagery are increasingly used for asset inspection and maintenance. These data enable detailed segmentation of critical components, such as power lines, pylons, and insulators, for updating geographic databases. However, current inspection workflows remain largely manual, making them both time-consuming and costly. Discussions with domain experts suggest that manual processing constitutes approximately one-third of the total acquisition cost. This challenge is further amplified by the anticipated growth in inspection volumes, driven by global efforts toward renewable energy integration and the corresponding expansion of electrical transmission networks.

Thus, multimodal fusion represents a promising approach to automate and enhance electrical infrastructure inspection workflows. Despite the growing need for multimodal approaches in infrastructure monitoring, research in this area remains hindered by the absence of publicly available datasets specifically designed for LiDAR-image fusion in electrical infrastructure segmentation. This lack of benchmark data limits the ability to develop, evaluate, and compare fusion techniques in operational applications.

To address this gap, we introduce \emph{GridNet-HD} (\textbf{Grid} electrical \textbf{Net}work with \textbf{H}igh \textbf{D}ensity LiDAR and high resolution images), a novel multimodal dataset specifically designed for 3D semantic segmentation of electrical infrastructures. \emph{GridNet-HD} consists of 7,694 high-resolution images and 2.5 billion precisely co-georeferenced LiDAR points, both annotated into 11 semantic classes. The dataset was collected across four distinct regions in Switzerland, covering mountainous, rural, and forest environments, ensuring diverse real-world conditions for robust model evaluation. Furthermore, electrical infrastructures follow highly standardized designs worldwide, meaning that models trained on \emph{GridNet-HD} have a strong potential for generalization to infrastructures in other countries, enhancing the dataset's utility for global applications. A detailed presentation of the dataset is provided in Section~\ref{sec:dataset}.

\emph{GridNet-HD} is the first publicly available multimodal dataset tailored specifically for electrical infrastructure applications. Its unique characteristics also make it highly relevant for general LiDAR-image fusion research:
\begin{itemize}
    \item High-density LiDAR: Captured at densities ranging from 200 to 800~pts/m².
    \item High-resolution imagery: Images with a ground sampling distance (GSD) of 1.5~cm.
    \item Precise co-registration: Robust alignment achieved through direct georeferencing, refined with aerotriangulation and Ground Control Points (GCPs).
    \item Manual semantic annotations: Both 2D and 3D segmentation annotations for powerline structures and environmental context.
    \item Long-term accessibility: Hosted online, ensuring sustainable and convenient access.
\end{itemize}

This contribution is in line with the future research vision proposed in the recent review~\cite{reveiew_elec_cvpr2025}, which emphasizes the lack and importance of publicly accessible datasets and calls for a greater focus on multimodal deep learning techniques to advance automated power line inspection.

\section{Related work} \label{sec:related_work}
\subsection{Summary of datasets}
The fusion of LiDAR and image data has been extensively studied in computer vision, notably for tasks such as semantic segmentation, object detection, and 3D reconstruction. Prominent examples include widely used benchmarks for autonomous driving~\cite{waymo,Caesar_2020_CVPR,kitti360} and indoor scene understanding~\cite{scannet++,ScanNet,stanford2d3ds}. In contrast, multimodal datasets specific to aerial remote sensing, particularly for electrical infrastructure monitoring, remain rare or non-existent.

Table~\ref{tab:powerline_datasets} summarizes publicly available multimodal LiDAR-image datasets across three application domains: autonomous driving, indoor scene understanding, and UAV (Unmanned Aerial Vehicle)-based environmental and infrastructure monitoring. The review emphasizes electrical infrastructures, for which an exhaustive coverage is attempted regardless of modality. Well-established datasets from autonomous driving and indoor environments are included for two reasons: (1) they constitue benchmarks that have shaped multimodal learning practices, particularly in LiDAR-image fusion; and (2) they underline the lack of comparably large-scale, high-resolution, and accurately co-registered datasets for power grid inspection. A selection of UAV-based multimodal datasets is also incorporated, primarily developed for other applications such as terrain mapping, forestry, or urban monitoring, yet serving as relevant points of comparison. Including these UAV datasets helps to  better contextualize the significance of \emph{GridNet-HD} as one of the largest UAV-acquired multimodal datasets specifically tailored for high-resolution 3D semantic segmentation of electrical infrastructures.

\begin{table*}[h]
    \centering
    \renewcommand{\arraystretch}{1.3}
    \resizebox{\textwidth}{!}{%
    \begin{tabular}{clcccccc}
        \toprule
        \textbf{Category} & \textbf{Dataset} & \textbf{Scene Type} & \textbf{Density / \# points} & \textbf{Image GSD / \# Images} & \textbf{\# Classes} & \textbf{Sensor} & \textbf{Annotations} \\ 
        \cmidrule(r){1-8}
        \multirow{3}{*}{\rotatebox[origin=c]{90}{\parbox{1.5cm}{\centering \textbf{A. D.}}}} & Waymo Open~\cite{waymo} & urban & - / $\sim$70~B & - / $\sim$390~k & 23 & MLS & 2D \& 3D S \\  
        & nuScenes~\cite{Caesar_2020_CVPR} & urban & - / $\sim$1.1~B & - / $\sim$1.4~M & 32 & MLS & 3D S \\ 
        & KITTI360~\cite{kitti360} & urban & - / $\sim$12~B & - / $\sim$320~k & 19 & MLS & 2D \& 3D S \\ 
        \cmidrule(r){1-8}
        
        \multirow{3}{*}{\rotatebox[origin=c]{90}{\parbox{1.6cm}{\centering \textbf{I. S. U.}}}}  
        & ScanNet~\cite{ScanNet} & indoor & - / 242~M & - / $\sim$2.5~M & 20 & RGB-D & 2D \& 3D S \\ 
        & S3DIS~\cite{stanford2d3ds} & indoor & - / 273~M & - / $\sim$70~k & 13 & Matterport & 2D \& 3D S \\ 
        & ScanNet++~\cite{scannet++} & indoor & (460 scenes) & - / $\sim$4~M & 1000 & RGB-D & 2D \& 3D S \\ 
        \cmidrule(r){1-8}
        \multirow{3}{*}{\rotatebox[origin=c]{90}{\parbox{1.5cm}{\centering \textbf{UAV}}}} & Individual tree~\cite{Individual_tree} & forest & 37 pts/m\textsuperscript{2} / 2~M  & 7~cm (orthophoto) & 7 & ALS & 2D B \\  
        &Hessigheim3D~\cite{hessigheim} & urban aerial &  800 pts/m\textsuperscript{2} / 74~M & LiDAR RGB colorized & 11 & ALS & 3D S \\ 
        &SensatUrban~\cite{Sensaturban} & urban aerial & - / 2.8~B & - & 13 & P & 3D S \\ 
        \cmidrule(r){2-8}
        \multirow{3}{*}{\rotatebox[origin=c]{90}{\parbox{7cm}{\centering \textbf{Electrical \\ Infrastructure}}}} & STN PLAD~\cite{stn_plad} & elec. &  - & -/133 & 5 & RGB & 2D B \\ 
        &INS PLAD~\cite{InsPLAD} & elec. &  - & - / 10,607 & 17 & RGB & 2D B \\ 
        &PLD UAV~\tablefootnote{https://github.com/SnorkerHeng/PLD-UAV} & elec. & -  & - / 860 & 1 & RGB & 2D B \\ 
        &PTLD~\cite{Power_transmission_line_dataset} & elec. & - & - / 348 (real), 696 (virtual) & 3 & RGB & 2D B \\ 
        &PowerLineImageDataset~\tablefootnote{https://data.mendeley.com/datasets/n6wrv4ry6v/8} & elec. & - & - / 4,000 (IR), 4,000 (RGB) & 2 & RGB & 2D B \\ 
        &PTLAI~\cite{ptlai} & elec. & -  & - / 6,295 & 5 & RGB & 2D B \\ 
        &TTPLA~\cite{ttpla}& elec. & - &  - / 1,100  & 2 & RGB &  2D B \\ 
        &DALES~\cite{dales}& urban aerial & 50~pts/m\textsuperscript{2} / 505~M &  -  & 8 (2 elec.) & ALS & 3D S  \\ 
        &ECLAIR~\cite{eclair2024}& elec. & 50~pts/m\textsuperscript{2} / 582~M  & LiDAR RGB colorized   & 11 & ALS & 3D S \\ 
        &CPLID~\cite{CPLID}& elec. & - & - / 848   & 1 & RGB & 2D B  \\ 
        &Tower dataset~\cite{Towerdataset}& elec. & - &  - / 1,300  & 1 & RGB & 2D B  \\ 
        &Tomaszewski et al.~\cite{TOMASZEWSKIdataset}& elec. & - &  - / 2,630  & 1 & RGB & 2D B \\ 
        &Power line dataset~\cite{powerlinedataset}& elec. & - &  - / 4,200  & 1 & RGB & 2D S  \\ 
        \cmidrule(r){2-8}
        &\textbf{GridNet-HD (ours)}& \textbf{elec.} & \textbf{200 to 800 pts/m\textsuperscript{2} / 2.5~B} & \textbf{1.5~cm / 7,694}  & \textbf{11} & ALS & \textbf{2D \& 3D S}  \\ 
        \bottomrule
    \end{tabular}
    }
    \caption{
    Overview of publicly available multimodal LiDAR-image datasets categorized by application domain: autonomous driving (A.D.), indoor scene understanding (I.S.U.), UAV-based mapping, and electrical infrastructure monitoring. While the table lists a selection of well-established datasets in the first three domains, it aims to provide an exhaustive survey of publicly available datasets for power grid inspection (whether or not it is multimodal). The datasets are compared by scene type, 3D point cloud density and number of points, image GSD (Ground Sampling Distance) and number of images, number of annotated classes, sensor type (MLS: Mobile Laser Scanning, ALS: Aerial Laser Scanning, P: Photogrammetry), and the type of annotations provided (S for semantic and B for boxes).}
    \label{tab:powerline_datasets}
\end{table*}

\subsection{Identified gaps}

\subsubsection{Autonomous driving}

Large-scale autonomous driving datasets significantly advanced multimodal LiDAR-image fusion research but are designed primarily for dynamic urban scenes, focusing on vehicles and pedestrians rather than static infrastructure (e.g., poles, cables, towers). Additionally, these datasets often lack precise image-LiDAR co-registration optimized by rigorous photogrammetric methods, leading to alignment issues especially for thin or distant objects critical in electrical infrastructure monitoring.

\subsubsection{Indoor scene understanding}
Indoor multimodal datasets provide rich semantic annotations for structured indoor environments. However, their applicability to outdoor infrastructure monitoring is limited by the constrained, structured nature of indoor scenes and their frequent reliance on RGB-D sensors rather than precise LiDAR acquisition, reducing their relevance for accurate aerial inspections.

\subsubsection{Electrical infrastructure monitoring}

Currently available electrical infrastructure datasets usually contain either LiDAR or imagery alone, without integrated multimodal annotations. This lack of multimodal, precisely co-registered data hinders the development and benchmarking of fusion methods specifically optimized for aerial inspection of electrical grids.

\subsection{Addressing the gap}

To address these limitations, \emph{GridNet-HD} is introduced as the first publicly available multimodal dataset specifically designed for 3D semantic segmentation of electrical infrastructure. \emph{GridNet-HD} provides:
\begin{itemize}
    \item High-density LiDAR (200–800 pts/m²) and high-resolution imagery (1.5 cm GSD);
    \item Precise multimodal co-registration, refined through aerotriangulation and Ground Control Points (GCPs);
    \item Detailed manual semantic annotations in both 2D and 3D for infrastructure elements and their surroundings.
\end{itemize}
    
Although smaller in total size compared to autonomous driving datasets, \emph{GridNet-HD} uniquely offers high spatial resolution and high co-registration accuracy, making it ideal for fine-grained electrical infrastructure inspection and potentially beneficial to broader tasks such as land cover mapping.

\section{Dataset description} \label{sec:dataset}

The \emph{GridNet-HD} dataset comprises 36 distinct areas captured at four different locations across the cantons of Vaud, Valais, and Fribourg in Switzerland, covering diverse landscapes such as mountainous terrains, plains, fields, and forests areas. Detailed maps illustrating these areas are provided in supplementary materials~\ref{app:map}. Each area corresponds to a unique drone flight, producing a dataset consisting of both a point cloud and a set of images. These datasets model segments of power lines ranging in length from 500 to 1300~m, with a typical cross-sectional width of 60 to 80~m. Small overlaps between certain areas were intentionally designed to ensure continuity along the power lines.

To generate this high-quality, densely annotated 2D and 3D dataset, a rigorous data acquisition and processing protocol were established, incorporating multiple checkpoints throughout various stages to ensure data integrity and accuracy. The following sections detail the procedures for acquisition, co-registration, and labeling of the 3D data, as well as the subsequent reprojection of these labels onto the corresponding 2D images.

\subsection{Dataset acquisition}

Data acquisition was performed using a DJI Matrice 350 RTK UAV equipped with a DJI Zenmuse L2 sensor, which integrates a LiDAR sensor and a CMOS optical camera.

Optical imagery was captured using the integrated Zenmuse L2 camera (resolution of 5280x3956~pixels) configured in an oblique orientation, ensuring comprehensive visibility of the lateral aspects of pylons and insulators. The Ground Sampling Distance (GSD) achieved at the pylon tops is 0.5 to 1~cm.

\begin{figure*}[htb]
  \centering
  \includegraphics[width=0.8\linewidth]{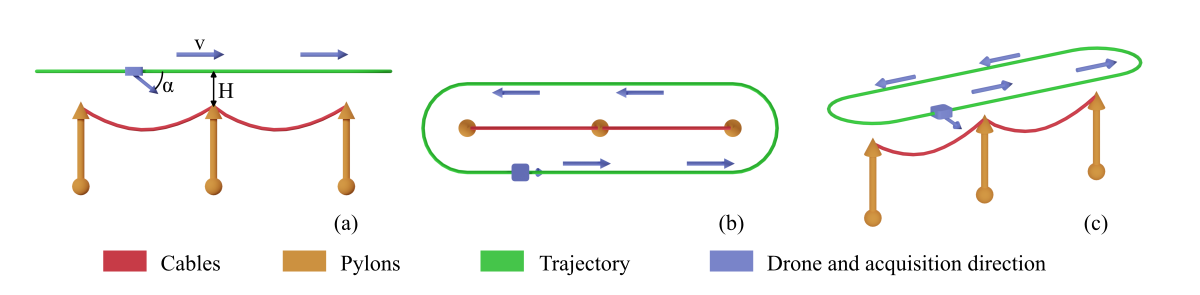}
  \caption{Schematic view of a flight plan in comparison with the power line. The angle $\alpha$ (here of 50°) of the oblique view and the height H over the power line are input parameters of the calculated trajectory. (a) side view, (b) top view, (c) perspective view.}
  \label{fig:trajectory}
\end{figure*}

The UAV trajectory follows the power line alignment directly, at a consistent altitude of 25~m above the power line. To optimize visual coverage of pylon structures, the Zenmuse L2 sensor was oriented obliquely at an angle $\alpha$ of 50° from the horizon. Acquisition flights were conducted bidirectionally along the power line, facilitating full coverage of both sides of the pylons, as illustrated in Figure~\ref{fig:trajectory}, and increasing the resulting point cloud density.

The UAV flight trajectory was automatically planned using pre-defined geospatial information, including the geographic coordinates, height, and elevation of each targeted pylon. These parameters enabled accurate localization of the pylon tops, which served as key reference points for trajectory generation. Assuming unobstructed airspace above the conductors, linear flight segments were computed between successive pylons. These segments were then concatenated to form the complete flight path, with a vertical offset of 25~m applied to ensure clearance above the power line. Each flight segment was executed in both directions, with a slight lateral offset applied. U-turn maneuvers were integrated at each trajectory endpoint to allow continuous data acquisition across the entire power line.

To balance acquisition efficiency with the point cloud density required for accurate modeling of small-scale structures, such as insulators, a variable-speed flight approach was adopted. This method involves decreasing the UAV speed near pylons to maximize data density, then increasing speed along sections between pylons. Flight speeds varied between 2 and 10~m/s. Due to the oblique orientation of the sensor, speed reduction was strategically initiated prior to reaching each pylon. This anticipatory adjustment was calculated using tower height data combined with the known sensor inclination angle.

\subsection{Dataset co-registration}

Georeferencing of the acquired data was performed using RTK GPS positioning combined with inertial measurements provided by the Zenmuse L2 sensor. The DJI Terra software was employed for the initial alignment and processing of LiDAR data, resulting in georeferenced point clouds in UTM 32N coordinate system (EPSG:32632).
LiDAR data alignment was manually verified through visual inspection in each acquisition zone, focusing specifically on potential misalignments between forward and backward flight paths. No significant misalignments exceeding a few centimeters were observed within the central, infrastructure-critical portions of the dataset. However, minor positional discrepancies, ranging between 5 and 10~cm, were occasionally identified in peripheral areas with high vegetation. Given their locations, these residual offsets were deemed acceptable and not detrimental to the overall quality or objectives of the dataset.

The resulting LiDAR point clouds were subsequently colorized using optical imagery captured simultaneously by the Zenmuse L2 camera. In addition to RGB values derived from images, the final point clouds contain various attributes, including intensity, echo number, total echoes, and scan angles. The achieved point cloud density ranges between 200 and 800~points/m\textsuperscript{2}, corresponding to a spacing of one point every 3 to 7~cm.

However, image georeferencing produced by DJI Terra exhibited unsatisfactory results, showing noticeable discrepancies relative to the LiDAR data. This limitation is primarily due to relying solely on direct georeferencing without the use of Ground Control Points (GCPs). Given that our objective was to ensure accurate relative co-registration between images and LiDAR, we opted to utilize the LiDAR data as our geometric reference. Consequently, the subsequent processing steps were dedicated to refine the alignment of images with the LiDAR reference.

\begin{figure}[htb]
  \centering
  \includegraphics[width=\linewidth]{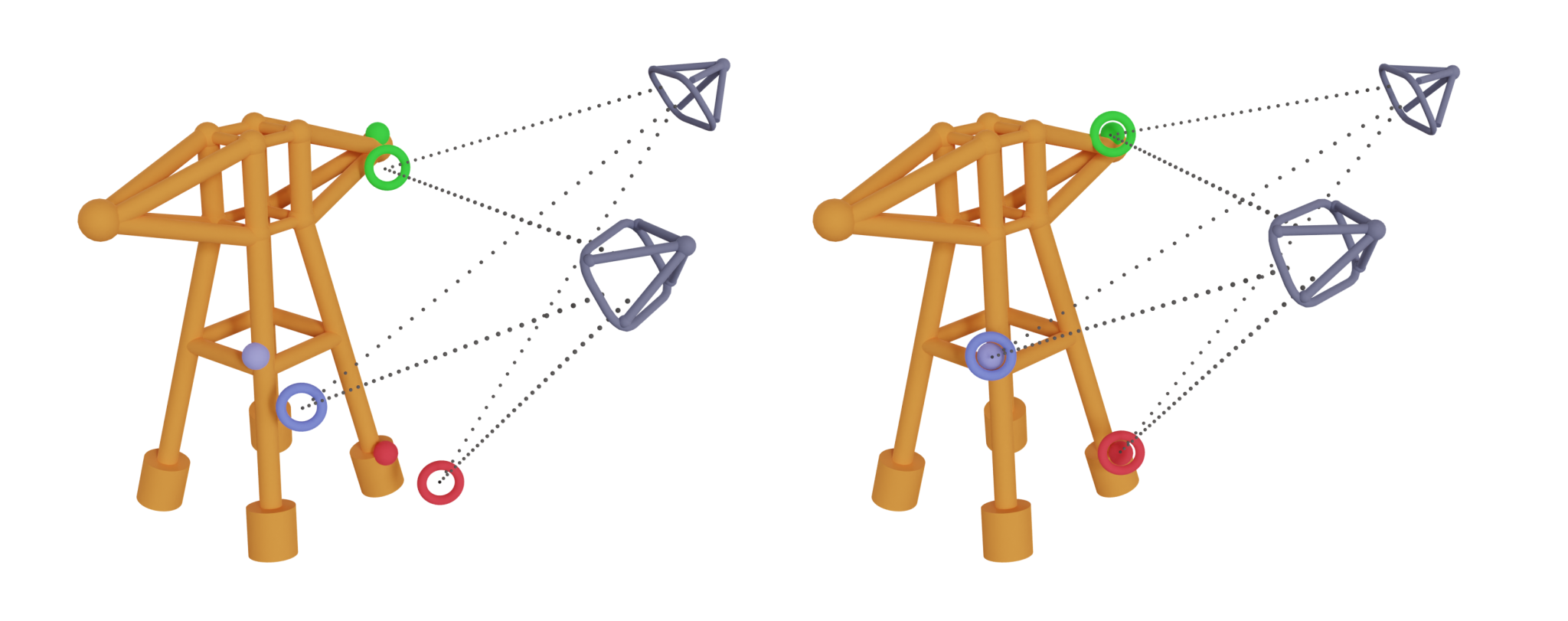}
  \caption{Manual GCP clicking for precise image–LiDAR co-registration. Left: before refinement, 3D points (filled dots) project away from the clicked image features (rings) across views. Right: after aerotriangulation using GCPs, projections align with image features, reducing residuals and enforcing consistent geometry.}
  \label{fig:virtual_gcps}
\end{figure}

Although existing semi-automatic approaches (e.g., multi-sensor fusion-based methods~\cite{kiriaky_open_journal}) are available for LiDAR-image registration, a manual procedure was selected. Initially, clearly identifiable reference points were visually selected in the LiDAR data and then precisely matched in the corresponding images. These reference points served as GCPs, with the LiDAR coordinates acting as the spatial reference (this step is illustrated on Figure~\ref{fig:virtual_gcps}).

After careful selection and spatial distribution of these LiDAR based GCPs throughout each survey area, a bundle adjustment was performed using Agisoft Metashape software. This adjustment employed a realistic stochastic model characterized by centimeter-level accuracy for both positional parameters of images and the coordinates of the GCPs, along with degree-level accuracy for angular orientations. Additionally, LiDAR data were integrated into the bundle adjustment process, providing supplementary geometric constraints.

To evaluate the accuracy of the co-registration between LiDAR and image data, residuals on GCPs were examined in image coordinates. Upon obtaining residuals consistently within a few centimeters, dense image-based point clouds were generated through photogrammetric dense correlation. 
Subsequently, cloud-to-cloud distance computations were performed between the photogrammetric point clouds and their corresponding LiDAR datasets.
Visual inspection confirmed that discrepancies remained within a few centimeters. If larger deviations were observed, the bundle adjustment was iteratively repeated, incorporating additional GCPs strategically placed in problematic areas to enhance overall alignment accuracy.

\begin{figure}[t]
    \centering
    \begin{subfigure}[b]{0.366\columnwidth}
        \includegraphics[width=\linewidth]{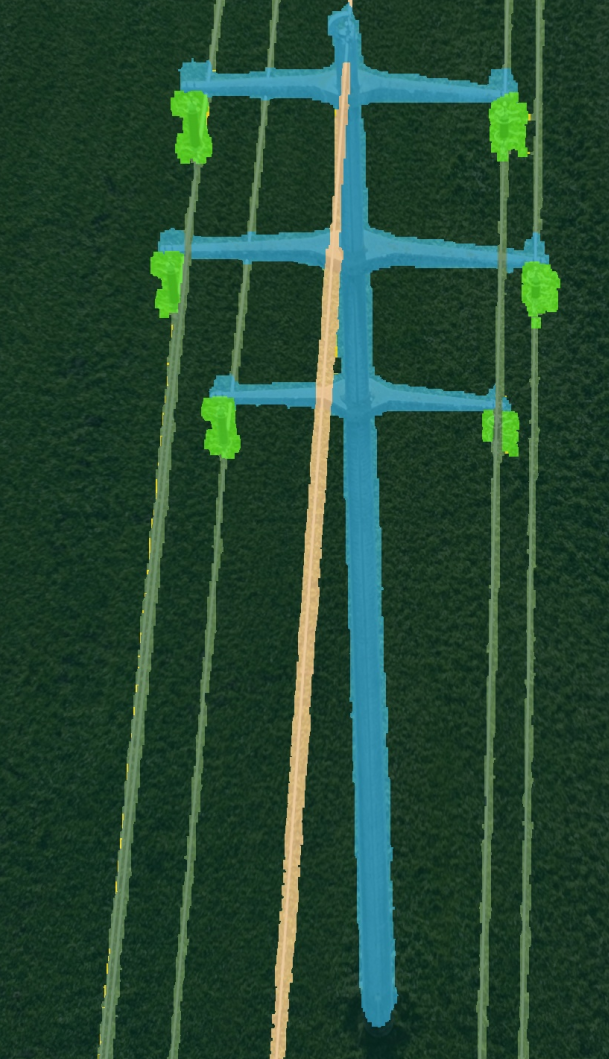}
        \caption{t5b5: 3D to 2D check.}
    \end{subfigure}
    \hfill
    \begin{subfigure}[b]{0.6\columnwidth}
        \includegraphics[width=\linewidth]{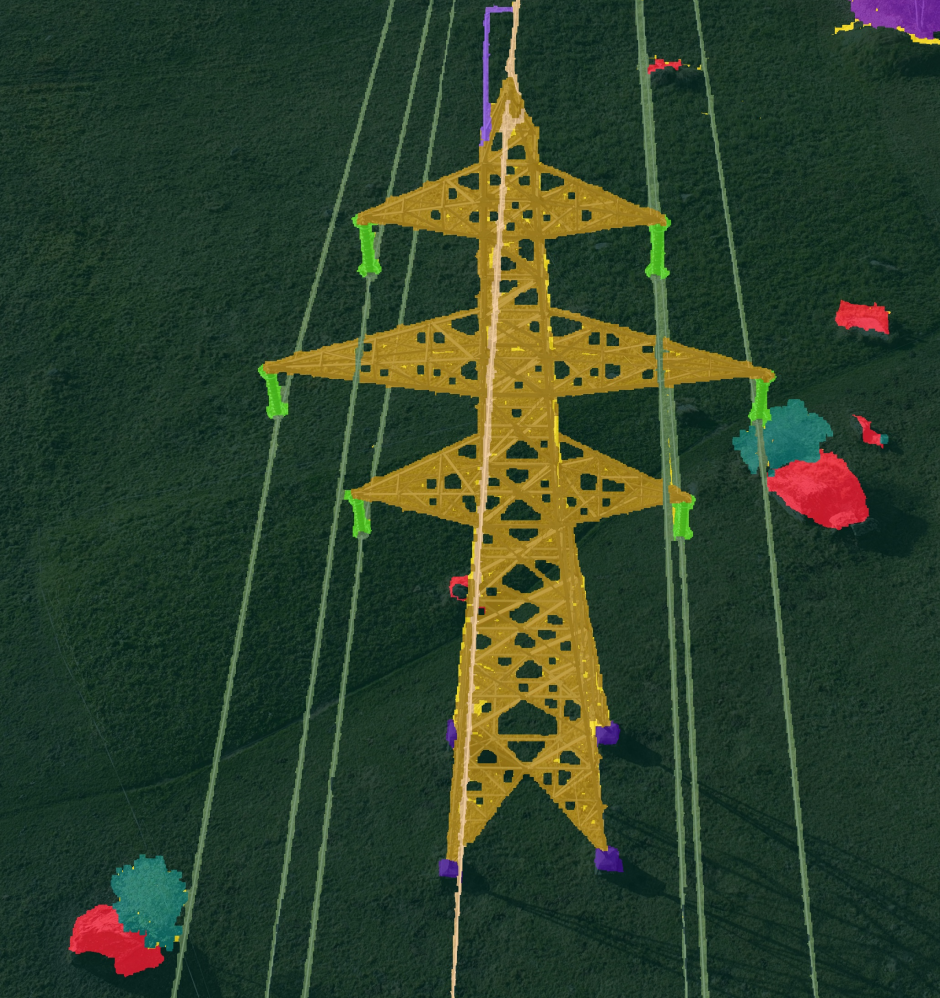}
        \caption{t6z4: 3D to 2D check.}
    \end{subfigure}
    \caption{Successful 3D to 2D label projection checks after final alignment refinement.}
    \label{fig:final_projection_check}
\end{figure}

As a final validation step, we assessed the projection quality of 3D semantic labels onto 2D images, semantic labeling process is detailed in Section~\ref{sec:label_3d_2d}. A random subset corresponding to approximately 25~\% of the images was manually reviewed to verify the visual alignment of the projected labels. If any misalignment was deemed visually unacceptable, an additional manual georeferencing step was performed by introducing new GCPs, particularly in the affected areas. This refinement process was repeated until a satisfactory projection quality was consistently achieved across the sampled images. Figure~\ref{fig:final_projection_check} illustrates two examples of satisfactory visual checks after the final refinement step.

\subsection{Dataset annotation}

\subsubsection{Semantic class definition}

The original \emph{GridNet-HD} dataset provides annotations across 22 semantic classes, along with an additional unassigned class. To address the significant class imbalance observed in some underrepresented categories, we introduce a semantic regrouping strategy that consolidates the 22 classes into 11 broader categories, while preserving the unassigned label (pylon, conductor cable, structural cable, insulator, high vegetation, low vegetation, herbaceous vegetation, rock/gravel, impervious soil, water and building). This grouping is designed to maintain semantic coherence and improve the robustness of model training, particularly for rare classes. A complete description of the original class set and the proposed regrouping is provided in supplementary materials~\ref{app:classes}.

\subsubsection{Manual labelling process}\label{sec:label_3d_2d}

The LiDAR point clouds were manually annotated by two coordinated operators to maintain the same rules using polygonal selection tools in CloudCompare. Semantic labels were directly assigned to 3D regions based on geometric and contextual information visible in the point clouds.
These 3D annotations were subsequently reprojected into the corresponding image views. This strategy is commonly adopted in multimodal datasets such as Waymo Open~\cite{waymo} and KITTI-360~\cite{kitti360}. This reprojection process explains why certain image regions remain unlabeled, as image annotations rely strictly on the visibility and coverage of the LiDAR data.

For this reprojection, the method described in~\cite{carreaud_igarss23, carreaud_rar_isprs} was adapted and enhanced through computational optimizations and the explicitly integration of depth maps to handle occlusions. The formalism for projecting a 3D point into image coordinates, as well as the computation of pixel-to-LiDAR visibility, is detailed in supplementary materials~\ref{app:proj}. Using depth maps allows us to avoid common reprojection artifacts, such as incorrectly projecting LiDAR ground points onto image pixels corresponding to obliquely viewed objects such as trees. Examples of the resulting LiDAR-to-image label reprojection are shown in Figure~\ref{fig:final_projection_check}.

\subsubsection{Train/test split}

The train/test split was carefully designed to ensure representative class distributions across the proposed 11-class grouping. In addition, we further propose a division of the training set into separate training and validation subsets. 
A brief summary of the split showing the number of geographic zones used for training/validation and testing, as well as global point statistics, is provided in Table~\ref{tab:mini_split}, while full per-class statistics remain available in supplementary materials~\ref{app:split}.

\begin{table}[h]
\small
\centering
\begin{tabular}{lccc}
\toprule
 & \textbf{Train/Val} & \textbf{Test} & \textbf{Total} \\
\midrule
\# Zones & 27 & 9 & 36 \\
Points (M) & 1,690 & 759 & 2,449 \\
Test / Total (\%) & -- & 31.0\% & -- \\
\bottomrule
\end{tabular}
\caption{Dataset split statistics.}
\label{tab:mini_split}
\end{table}

\section{Baseline methods}
\label{sec:baseline}

A set of baseline models is evaluated on the \emph{GridNet-HD} dataset to establish reference performance.. 
For the baselines with fast training cycles, namely the \emph{ImageVote}, \emph{SPT}, and \emph{MLP (late fusion)}, we train each model three times with different random seeds. For all other baselines, only a single training run is performed due to their higher computational cost.
Table~\ref{tab:lidar_image_fusion_results} reports the best single-run mean Intersection over Union (mIoU) on the test set for each method. Complete per-class IoUs and, for the multi-run baselines, the mean and standard deviation across seeds are provided in supplementary materials~\ref{app:baseline_results}. 

Implementations and \emph{GridNet-HD} configurations for all baselines are available here: \url{https://huggingface.co/collections/heig-vd-geo/gridnet-hd}.

\subsection{ImageVote (image-only reprojection)}
\label{sec:imagevote}

\textbf{Overview.}
This baseline transfers 2D semantic predictions to 3D points without any 3D supervision.
A transformer-based 2D segmenter (UPerNet~\cite{upernet} with a Swin-Tiny backbone~\cite{swin}) is first applied to each RGB image to produce per-pixel class logits.

Given calibrated cameras with intrinsics $K_c$ and extrinsics $(R_c, \mathbf{t}_c)$, each LiDAR point $\mathbf{p}\!\in\!\mathbb{R}^3$ is reprojected into all views $c\in\mathcal{C}$ via
\(
\mathbf{x}_c \;=\; \pi\!\big( K_c\,[\,R_c \mid \mathbf{t}_c\,]\, \mathbf{p}).
\)
This projection gives image coordinates $\mathbf{x}_c$ when inside the field of view (we used the professional software Metashape and coded all the exact reprojections in an optimized way to be independent of any license. The Metashape formalism is detailed in the supplemental material~\ref{app:proj}).

\textbf{Visibility filtering.}
To mitigate errors caused by occlusions and minor pose misalignments, depth-consistency checks are applied.
Let $z_{\text{lidar}}^{(c)}(\mathbf{p})$ be the depth of $\mathbf{p}$ along camera $c$’s optical axis and $z_{\text{img}}^{(c)}(\mathbf{x}_c)$ the depth rendered from depth map.
The projection is retained only if
\(
\big|\,z_{\text{img}}^{(c)}(\mathbf{x}_c) - z_{\text{lidar}}^{(c)}(\mathbf{p})\,\big| \le \tau_z.
\)
with \(\tau_z\) a fixed threshold. Rejected views do not contribute to the point’s label.

\textbf{Logit aggregation.}
For each valid view, the per-class softmax-logits $\boldsymbol{\ell}_c(\mathbf{x}_c)\in\mathbb{R}^K$ at pixel $\mathbf{x}_c$ are read and aggregated across views:
\(
\mathbf{L}(\mathbf{p}) \;=\; \sum_{c \in \mathcal{V}(\mathbf{p})} w_c \,\boldsymbol{\ell}_c(\mathbf{x}_c),
\)
where $\mathcal{V}(\mathbf{p})$ is the set of visibility-filtered views, and $w$ the weighting applied to each view (default is 1 but may be inverse to the camera-point distance or linked to the evaluation of optimal viewing conditions).

\textbf{Relation to prior work.}
The proposed approach performs a direct, projection-based transfer of 2D semantics to 3D, similar to methods that supervise 3D with 2D signals~\cite{genova_learning3dfrom2d,Reichardt_360to3d}.
CLIP-based pipelines~\cite{clipfo3d,partslip} leverage vision-language pretraining to bridge 2D-3D without dense 3D labels, while SAM3D~\cite{SAM3d} uses Segment Anything~\cite{sam} masks fused via a semantic NeRF~\cite{semantic_nerf}.
In contrast, the proposed method relies solely on explicit photogrammetric reprojection and per-pixel logits, without volumetric representations or learned 2D-3D feature alignment.
Comparable projection-driven 3D segmentation schemes have been reported across applications~\cite{Beniaouf2021,Pellis2022,Pellis2025}, including electrical infrastructure inspection~\cite{maj_vot_elec_deeplearning_pc}.

\subsection{3D-Only Baselines: SuperPoint and Point Transformers}
\label{sec:3donly}

\textbf{SuperPoint Transformer (SPT).}
SPT~\cite{robert2023spt}, a lightweight graph-based model operating on \emph{superpoints} (geometrically homogeneous clusters) is adopted.
To inject appearance cues, LiDAR points are optionally decorated with per-point RGB projected from imagery, resulting in a form of early fusion, but not strictly so, because the richness of the image information is lost. As also observed in~\cite{deepviewagg}, decorating points with per-point RGB alone typically underperforms true image–point fusion, which explicitly combines cues across modalities.

\textbf{Point Transformer v3 (PTv3).}
PTv3~\cite{PTv3}, a state-of-the-art backbone for large-scale point-cloud segmentation, is also included. In this setup, PTv3 operates on XYZ coordinates with per-point RGB, mirroring the SPT baseline. Configurations with voxel sizes of $5\,\mathrm{cm}$ ($\approx\!400$ pts/m$^2$) and $10\,\mathrm{cm}$ ($\approx\!100$ pts/m$^2$) are evaluated to analyze density–accuracy trade-offs, and results are reported both with and without test-time augmentation (TTA).

\textbf{Goal.}
These 3D-only baselines quantify the headroom of pure point-cloud methods and allow for isolating the added value of explicit image–LiDAR fusion.

\textit{Compared to 3D pipelines used directly in electrical-infrastructure studies~\cite{deeplearning_inspection_pc,electrical_inf_3d}, our baselines leverage recent state-of-the-art 3D segmentation architectures (SPT, PTv3), providing stronger 3D-only baseline references.}

\subsection{Fusion Baselines: Late Logit Fusion and DITR}
\label{sec:fusion_baselines}

\textbf{Late logit fusion (simple fusion)}
A minimal late-fusion strategy is implemented, in which, for each 3D point, the softmax logits from the image-only model (Sec.~\ref{sec:imagevote}) and the 3D SPT baseline (Sec.~\ref{sec:3donly}) are concatenated and passed through a lightweight MLP to predict the final label.
This follows late-fusion evidence from ImVoteNet~\cite{ImVoteNet} and MSeg3D~\cite{MSeg3D}, and aligns with LVIC~\cite{LVIC}, which shows that a very simple fusion can compete with more complex schemes provided that the pixel alignment is accurate.

\textbf{DITR ("DINO in the Room")}
The state-of-the-art DITR~\cite{DITR} is also included, This method extracts patch-level embeddings from a strong 2D foundation model (DINOv2~\cite{dinov2}) on the images, reprojects them into 3D space, and integrates them into a 3D segmentation backbone (Point Transformer v3~\cite{PTv3}).
The original architecture is followed and adapted to the \emph{GridNet-HD} setting. DITR serves as the strongest image+point fusion baseline in our experiments.

\begin{figure*}[t]
    \centering
    \begin{subfigure}[b]{0.24\textwidth}\includegraphics[width=\linewidth]{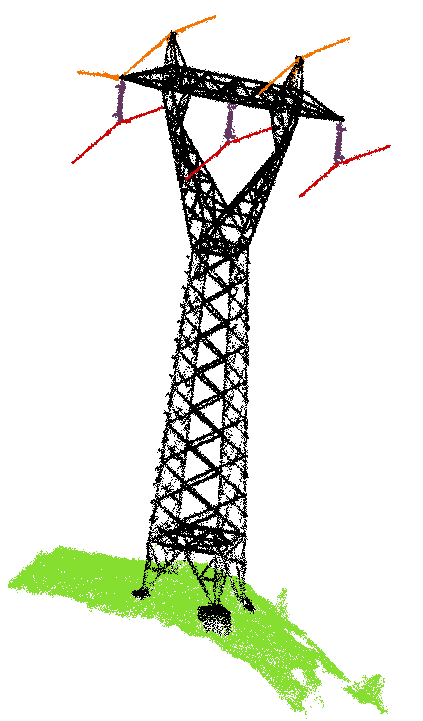}\caption{Ground truth}\end{subfigure}\hfill
    \begin{subfigure}[b]{0.24\textwidth}\includegraphics[width=\linewidth]{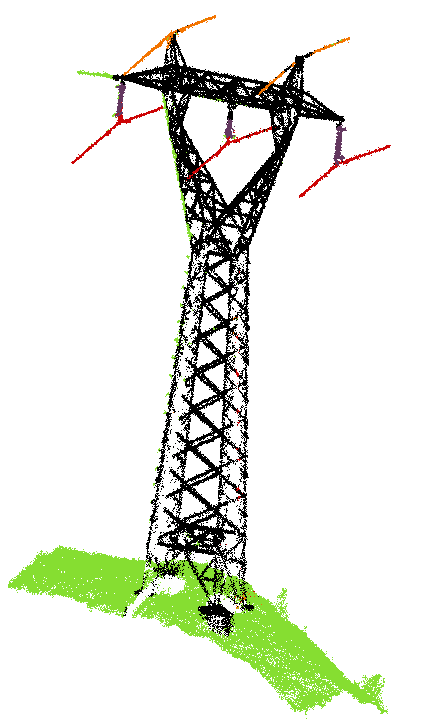}\caption{ImageVote}\end{subfigure}\hfill
    \begin{subfigure}[b]{0.24\textwidth}\includegraphics[width=\linewidth]{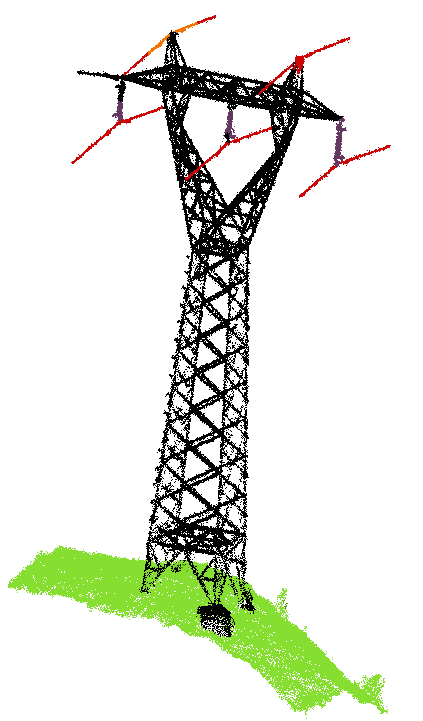}\caption{PTv3}\end{subfigure}\hfill
    \begin{subfigure}[b]{0.24\textwidth}\includegraphics[width=\linewidth]{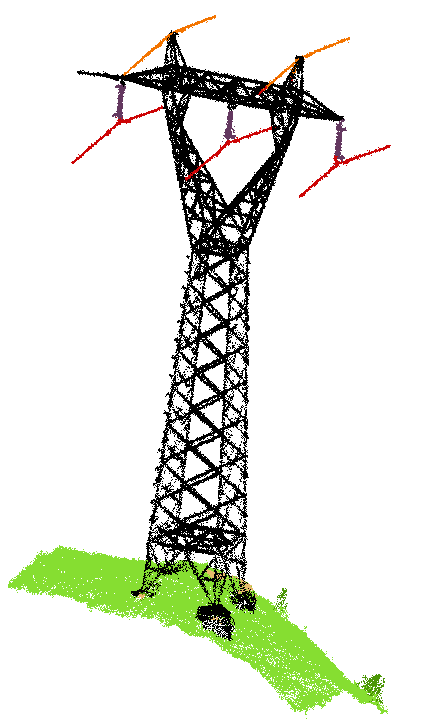}\caption{DITR}\end{subfigure}
    \caption{Qualitative comparison across three baselines on \emph{GridNet-HD}. From left to right: (a) Ground Truth; (b) ImageVote shows small projection issue and occlusion artifacts near object boundaries; (c) PTv3 (XYZ+RGB) yields smoother regions but misses thin structures (cables/insulators); (d) DITR produces the cleanest boundaries.}
    \label{fig:qualitative_analysis}
\end{figure*}

\subsection{Baseline comparison}
Table~\ref{tab:lidar_image_fusion_results} reports the main results on \emph{GridNet-HD} (per-class IoUs and variance over three runs are available in detailed results in supplementary material~\ref{app:baseline_results}). Three key findings are highlighted:

\textbf{(1) Higher point density helps.}
Within PTv3, using a finer voxel (\mbox{$5$\,cm} $\approx\!400$~pts/m$^2$) improves mIoU over \mbox{$10$\,cm} ($\approx\!100$~pts/m$^2$) (\mbox{$66.86$ vs.\ $64.53$}, $+2.33$), and test-time augmentation further raises PTv3 to \mbox{$69.32$}. This confirms that denser sampling benefits large-scale 3D segmentation in our setting.

\textbf{(2) Images add substantially more than per-point RGB.}

Per-point RGB decoration (e.g., SPT, PTv3) underperforms true image–point fusion; our fusion baselines clearly lead: Late Fusion MLP (\mbox{$74.22\%$}) and DITR with overlap\,+\,TTA (\mbox{\textbf{74.87\%}}).
This is consistent with prior work~\cite{deepviewagg}, which shows that explicit image–point fusion surpasses XYZ+RGB decoration by exploiting richer texture and spatial context.

\textbf{(3) Alignment outweighs fusion complexity (LVIC).}
Consistent with LVIC~\cite{LVIC}, performance depends mainly on the reliability of the image-LiDAR alignment: when pixel-point correspondences are accurate, a minimalist late logit fusion already yields strong results (\mbox{74.22\%} mIoU), surpassing either modality alone (\mbox{69.10\%}/\mbox{69.32\%}); with similarly good alignment, a SOTA fusion (DITR) provides a further but smaller gain (up to \mbox{74.87\%}), indicating that alignment matters more than fusion complexity.

Figure~\ref{fig:qualitative_analysis} illustrates these trends qualitatively: DITR delivers sharper tower/insulator boundaries, while PTv3 occasionally blurs class borders, and ImageVote exhibits projection artifacts near occlusions, which corresponds to our quantitative results.

\begin{table}[t]
\centering
\setlength{\tabcolsep}{10pt}
\scriptsize
\begin{tabular}{lcc}
\toprule
\textbf{Model} & \textbf{mIoU (\%)} & \textbf{Params} \\
\midrule
\multicolumn{3}{c}{\textit{Image-based}} \\
\midrule
ImageVote (2D $\rightarrow$ 3D reprojection) & \textbf{69.10} & 60M \\
\midrule
\multicolumn{3}{c}{\textit{3D-based}} \\
\midrule
SPT (XYZ+RGB) & 66.90 & 0.21M \\
PTv3 (XYZ+RGB, 10\,cm) & 64.53 & 46.2M \\
PTv3 (XYZ+RGB, 5\,cm) & 66.86 & 46.2M \\
\quad + TTA & \textbf{69.32} & 46.2M \\
\midrule
\multicolumn{3}{c}{\textit{Fusion (2D + 3D)}} \\
\midrule
Late Fusion MLP & 74.22 & 60.3M\textsuperscript{$\dagger$} \\
DITR (w/o overlap, w/o TTA) & 69.36 & 46.7M\textsuperscript{$\ddagger$} \\
DITR + overlap + TTA & \textbf{74.87} & 46.7M\textsuperscript{$\ddagger$} \\
\bottomrule
\end{tabular}
\vspace{1mm}
\begin{flushleft}
\footnotesize
\textsuperscript{$\dagger$}MLP adds only 78k parameters on top of ImageVote and SPT.\\
\textsuperscript{$\ddagger$}Parameters of frozen DINOv2-L not counted in training.
\end{flushleft}
\vspace{-0.4cm}
\caption{\textbf{Baseline comparison of 3D segmentation on the \emph{GridNet-HD} test set.}
Fusion approaches clearly outperform image-only and 3D-only methods. Best score per group in bold. Systematic use of overlap between batches in PTv3, TTA=Test Time Augmentations.}
\label{tab:lidar_image_fusion_results}
\end{table}

\section{Conclusion and Limitations}\label{sec:limitations}

\emph{GridNet-HD}, a new large-scale open dataset for image-LiDAR segmentation, was introduced to fill a major gap by providing high-density LiDAR (200-800 pts/m$^2$), high-resolution imagery (GSD 1.5\,cm), precise co-registration via direct georeferencing refined with aerotriangulation and GCPs. The dataset includes manual 2D/3D semantic annotations of powerline structures and thier surrounding environment. In addition, image-, 3D-, and fusion-based baselines were established, demonstrating that (i) higher point density (5\,cm vs.\ 10\,cm voxel) improves performance, (ii) explicit image-point fusion clearly outperforms XYZ+RGB decoration, and (iii) alignment quality matters more than fusion complexity.

External validation shows that models trained on \emph{GridNet-HD} yield visually compelling results suggesting strong generalization, though these samples cannot be released due to industrial constraints. Despite strong benchmark performance, several practical aspects remain.
Electrical components are globally standardised with low intra-class variability, favoring transfer across regions. Natural classes vary more and may require domain adaptation in ecologically distinct areas.
Effective fusion depends on accurate image–LiDAR calibration; \emph{GridNet-HD} reflects typical industry specifications, but misalignment can still harm performance. Class imbalance persists for fine-grained infrastructure classes (e.g., insulators, cables), though standard mitigation strategies (re-weighting, focal losses, targeted augmentation) remain easy to adopt.

\section{Acknowledgments}
We thank Jessica Bader for her significant effort in labeling a large portion of the dataset. We also acknowledge the experts from the AutoInspect3D project for their valuable feedback, expertise, and support, in particular: Yann Le Cahain and Bernard Valluy (Alpiq), Jean-Philippe Eberst and Nicolas Ackermann (CFF), David Ulrich and Julien Vallet (Helimap), Mokhtar Bozorg (HEIG-VD), Maximin Bron and Philipp Schaer (Orbis360), Daniel Gnerre (SIT de Vevey), Fabio Mariani and Patrice Poirier (SI de Genève), Adrian Gruenenfelder (SwissGRID), and Matthew Parkan and Marc Riedo (SIT de Neuchâtel). We are grateful to Xavier Muth and Fabien Délèze for the careful review, and to Jens Ingensand for his continued support. We also appreciate the insights provided by Laurent Jospin and the encouragement from the ESO laboratory. Finally, we thank the INSIT group at HEIG-VD for their constant support throughout this work.

The AutoInspect3D project was supported by a Young Researcher Grant from the HES-SO (University of Applied Sciences and Arts Western Switzerland), whose financial support is gratefully acknowledged.

{
    \small
    \bibliographystyle{ieeenat_fullname}
    \bibliography{main}
}

\clearpage
\setcounter{page}{1}
\maketitlesupplementary

\section{Acquisition zone map}\label{app:map}

\begin{strip}
\raggedright
\justifying

\begin{center}
\captionsetup{type=figure}
\includegraphics[width=0.69\linewidth]{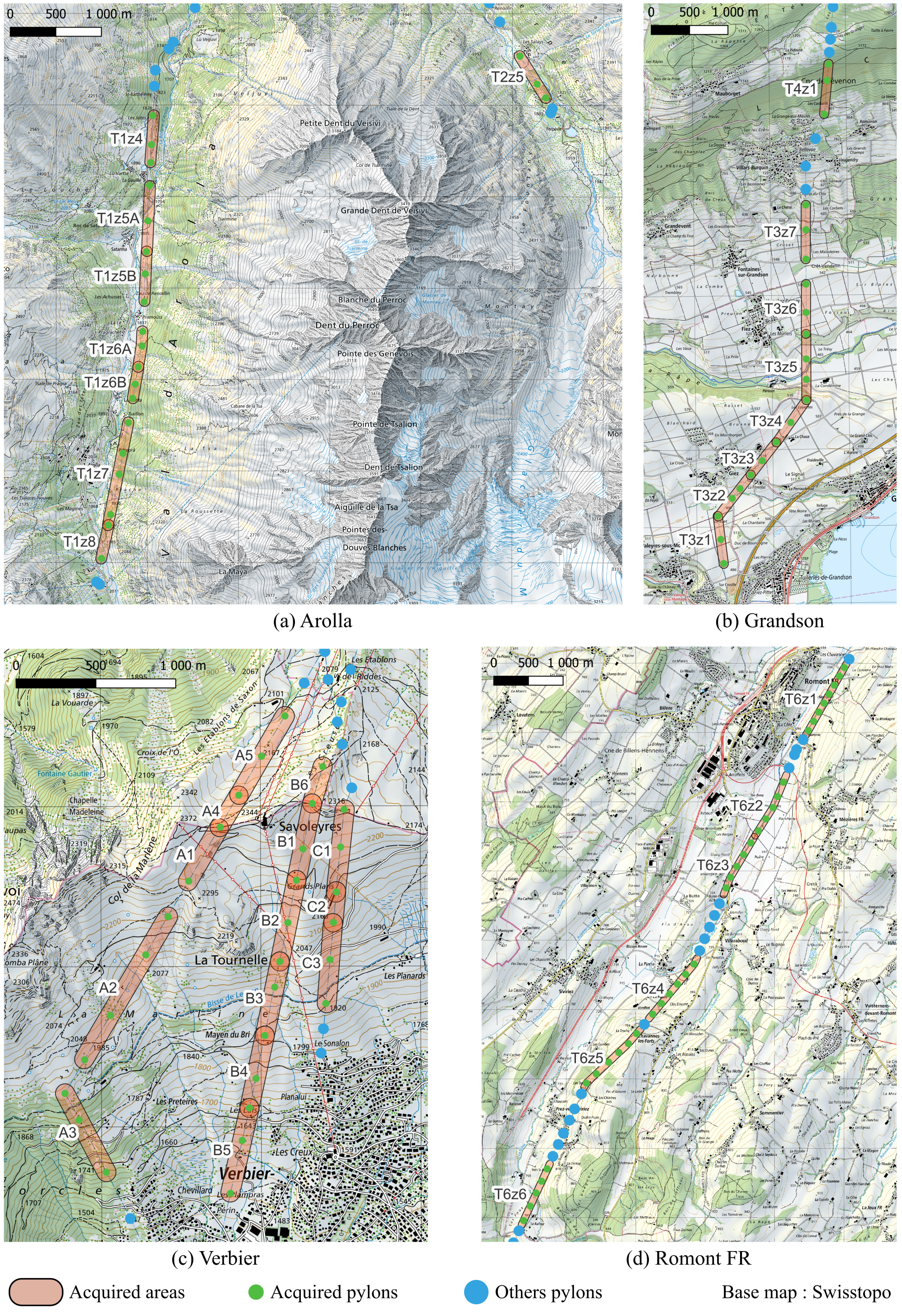}
\caption{Geographic distribution of the acquisition zones included in the \emph{GridNet-HD} dataset. Each zone corresponds to a specific area captured by UAV.}
\label{fig:dataset_zones}
\end{center}

\section{Original and remapped semantic classes}\label{app:classes}

\definecolor{pylonblue}{RGB}{165, 195, 244}
\definecolor{cableorange}{RGB}{249, 204, 158}
\definecolor{insulatorred}{RGB}{234, 154, 155}
\definecolor{vegdarkgreen}{RGB}{82, 140, 82}
\definecolor{veggreen}{RGB}{110, 162, 88}
\definecolor{veglightgreen}{RGB}{183, 216, 170}
\definecolor{soilgray}{RGB}{183, 183, 183}
\definecolor{roadbrown}{RGB}{231, 185, 177}
\definecolor{waterblue}{RGB}{160, 198, 232}
\definecolor{buildingbrown}{RGB}{167, 98, 98}
\definecolor{unassignedgray}{RGB}{65, 65, 65}

\begin{center}
\captionsetup{type=table}
\small
\begin{tabular}{cllc}
    \toprule
    \textbf{ID} & \textbf{Original Classes} & \textbf{Training Classes}                    & \textbf{ID}          \\ \midrule
    0           & Pylon foundation          & \cellcolor{pylonblue}                        & \cellcolor{pylonblue} \\
    1           & Cat head type pylon       & \cellcolor{pylonblue}                        & \cellcolor{pylonblue} \\
    2           & Triangle-arm pylon        & \cellcolor{pylonblue}                        & \cellcolor{pylonblue} \\
    3           & Portal pylon              & \cellcolor{pylonblue}                        & \cellcolor{pylonblue} \\
    4           & Other pylon               & \cellcolor{pylonblue}\multirow{-5}{*}{Pylon} & \cellcolor{pylonblue}\multirow{-5}{*}{0}                     \\ \midrule
    5           & Conductor cable           & \cellcolor{cableorange}Conductor cable & \cellcolor{cableorange} 1\\ \midrule
    6           & Guard cable               & \cellcolor{cableorange}     & \cellcolor{cableorange}   \\
    7           & Anchor cable              & \cellcolor{cableorange}\multirow{-2}{*}{Structural cable} & \cellcolor{cableorange}\multirow{-2}{*}{2} \\ \midrule
    8           & Suspension insulator - glass            & \cellcolor{insulatorred} & \cellcolor{insulatorred}  \\
    9           & Strain insulator - glass          & \cellcolor{insulatorred} & \cellcolor{insulatorred}  \\
    10          & Suspension insulator - porcelain              & \cellcolor{insulatorred} & \cellcolor{insulatorred}  \\
    11          & Strain insulator - porcelain            & \cellcolor{insulatorred}\multirow{-4}{*}{Insulator} & \cellcolor{insulatorred}\multirow{-4}{*}{3} \\ \midrule
    14          & High vegetation           & \cellcolor{vegdarkgreen}High vegetation & \cellcolor{vegdarkgreen} 4 \\ \midrule
    15          & Low vegetation            & \cellcolor{veggreen}Low vegetation & \cellcolor{veggreen}5 \\ \midrule
    16          & Herbaceous vegetation     & \cellcolor{veglightgreen}Herbaceous vegetation & \cellcolor{veglightgreen}6 \\ \midrule
    17          & Rock                      & \cellcolor{soilgray}   & \cellcolor{soilgray} \\
    18          & Gravel, soil              & \cellcolor{soilgray}\multirow{-2}{*}{Rock, gravel, soil} & \cellcolor{soilgray}\multirow{-2}{*}{7} \\ \midrule
    19          & Impervious soil (Road)    & \cellcolor{roadbrown}Impervious soil (Road) & \cellcolor{roadbrown}8 \\ \midrule
    20          & Water                     & \cellcolor{waterblue}Water & \cellcolor{waterblue}9 \\ \midrule
    21          & Building                  & \cellcolor{buildingbrown}Building & \cellcolor{buildingbrown}10 \\ \midrule
    12          & Other insulator           & \cellcolor{unassignedgray} & \cellcolor{unassignedgray}\\
    13          & Signage                   & \cellcolor{unassignedgray} & \cellcolor{unassignedgray}\\
    255         & Unlabeled                 & \cellcolor{unassignedgray}\multirow{-3}{*}{\textcolor{white}{Unassigned-Unlabeled}}  & \cellcolor{unassignedgray}\multirow{-3}{*}{\textcolor{white}{255}} \\ 
    \bottomrule
\end{tabular}
\caption{Mapping between the original 22 semantic classes and the 11 grouped classes used for training. Class grouping was designed to improve balance across categories while preserving semantic consistency.}
\end{center}

\section{Projection equations}\label{app:proj}

In addition to providing annotated multimodal data, \emph{GridNet-HD} includes optimized Python code for reprojecting 3D points into image space and computing depth maps. Although our initial implementation utilized formalisms from the commercial photogrammetry software Agisoft Metashape, we have independently coded all projection equations and depth map computations. This ensures that the tools we provide can be freely used without the need for paid licenses. All codes are published on HuggingFace: \url{https://huggingface.co/heig-vd-geo/ImageVote_GridNet-HD_baseline}.

The reprojection of 3D points onto the 2D image plane involves several transformations and corrections detailed below.

\paragraph{Rotation matrices.}
We define rotation matrices based on Euler angles $(\omega, \phi, \kappa)$ from Agisoft Metashape convention:

\begin{equation}
R_x(\omega)=\begin{bmatrix}
1 & 0 & 0 \\
0 & \cos\omega & \sin\omega \\
0 & -\sin\omega & \cos\omega
\end{bmatrix}\quad
\end{equation}

\begin{equation}
R_y(\phi)=\begin{bmatrix}
\cos\phi & 0 & -\sin\phi \\
0 & 1 & 0 \\
\sin\phi & 0 & \cos\phi
\end{bmatrix}\quad
\end{equation}

\begin{equation}
R_z(\kappa)=\begin{bmatrix}
\cos\kappa & \sin\kappa & 0 \\
-\sin\kappa & \cos\kappa & 0 \\
0 & 0 & 1
\end{bmatrix}
\end{equation}

The combined rotation matrix is:

\begin{equation}
R = R_z(\kappa) \cdot R_y(\phi) \cdot R_x(\omega)
\end{equation}

\paragraph{World to camera coordinates.}
A world coordinate point $M=\begin{bmatrix}X & Y & Z\end{bmatrix}^T$ is transformed into camera coordinates as:

\begin{equation}
\begin{bmatrix} X_c \ Y_c \ Z_c \end{bmatrix} = R (M - S),
\end{equation}

where $S=\begin{bmatrix}X_s & Y_s & Z_s\end{bmatrix}^T$ is the camera position.

\paragraph{Projection onto image plane.}
Normalized image coordinates $(x,y)$ and depth $z$ are computed:

\begin{equation}
x = -\frac{X_c}{Z_c}, \quad y = -\frac{Y_c}{Z_c}, \quad z = -Z_c
\end{equation}

\paragraph{Distortion corrections (Agisoft model).}
Radial and tangential distortions are corrected with coefficients $(k_1, k_2, k_3, k_4, k_5, p_1, p_2, p_3, p_4)$:

\begin{align}
r_c &= x^2 + y^2 \\
d_r &= 1 + k_1 r_c + k_2 r_c^2 + k_3 r_c^3 + k_4 r_c^4 + k_5 r_c^5 \\
d_{tx} &= p_1(r_c + 2x^2) + 2p_2xy(1+p_3 r_c + p_4 r_c^2) \\
d_{ty} &= p_2(r_c + 2y^2) + 2p_1xy(1+p_3 r_c + p_4 r_c^2) \\
x' &= x d_r + d_{tx}, \quad y' = y d_r + d_{ty}
\end{align}

Final pixel coordinates $(f_x, f_y)$ on the image plane are:

\begin{align}
f_x &= \frac{\text{width}}{2} + c_x + x' f + x' b_1 + y' b_2 \\
f_y &= \frac{\text{height}}{2} + c_y + y' f
\end{align}

where $f$ is the focal length, $(c_x,c_y)$ the principal point offsets, and $(b_1,b_2)$ sensor skew coefficients.

\paragraph{Depth map computation.}
Depth maps are computed by assigning the minimal depth value within a buffer around each projected pixel:

\begin{equation}
\text{depthmap}(i,j) = \min\left(\text{depthmap}(i,j), z\right)
\end{equation}

Visibility is determined by:

\begin{equation}
\text{visibility}(i,j) = \left|\text{depthmap}(i,j)-z\right| \leq \text{threshold}
\end{equation}

\section{Dataset splits and class distributions}
\label{app:split}

\subsection{Train/Test split}

The train/test split of the \emph{GridNet-HD} dataset was designed to preserve the overall semantic class distribution as defined in the 12-class grouping (11 semantic classes + 1 unassigned). Table~\ref{tab:split_test} reports the total number of points per class in the training and test sets, along with the percentage of points in the test set relative to the total, and the class-wise distribution within each subset. Figure~\ref{fig:split_visualization} visually illustrates the distribution of classes across the train and test sets.

\begin{center}
\small
\captionsetup{type=table}
\begin{tabular}{crrrccc}
%\toprule
\toprule
\textbf{Group ID} & \textbf{Train Points} & \textbf{Test Points} & \textbf{Total Points} & \textbf{\% Test/Total} & \textbf{Train Dist. (\%)} & \textbf{Test Dist. (\%)} \\ 
\midrule
0  & 11,490,104   & 3,859,573   & 15,349,677   & 25.1 & 0.7 & 0.5 \\
1  & 7,273,270    & 3,223,720   & 10,496,990   & 30.7 & 0.4 & 0.4 \\
2  & 1,811,422    & 903,089     & 2,714,511    & 33.3 & 0.1 & 0.1 \\
3  & 821,712      & 230,219     & 1,051,931    & 21.9 & 0.05 & 0.03 \\
4  & 278,527,781  & 135,808,699 & 414,336,480  & 32.8 & 16.5 & 17.9 \\
5  & 78,101,152   & 37,886,731  & 115,987,883  & 32.7 & 4.6 & 5.0 \\
6  & 1,155,217,319& 461,212,378 & 1,616,429,697& 28.5 & 68.4 & 60.7 \\
7  & 135,026,058  & 99,817,139  & 234,843,197  & 42.5 & 8.0 & 13.1 \\
8  & 13,205,411   & 12,945,414  & 26,150,825   & 49.5 & 0.8 & 1.7 \\
9  & 1,807,216    & 1,227,892   & 3,035,108    & 40.5 & 0.1 & 0.2 \\
10 & 6,259,260    & 2,107,391   & 8,366,651    & 25.2 & 0.4 & 0.3 \\
\midrule
\textbf{TOTAL} & 1,689,540,705 & 759,222,245 & 2,448,762,950 & 31.0 & 100 & 100 \\
\bottomrule
\end{tabular}
\caption{Train/test split statistics per semantic class.}
\label{tab:split_test}
\end{center}

\begin{center}
\captionsetup{type=figure}
\includegraphics[width=1\textwidth]{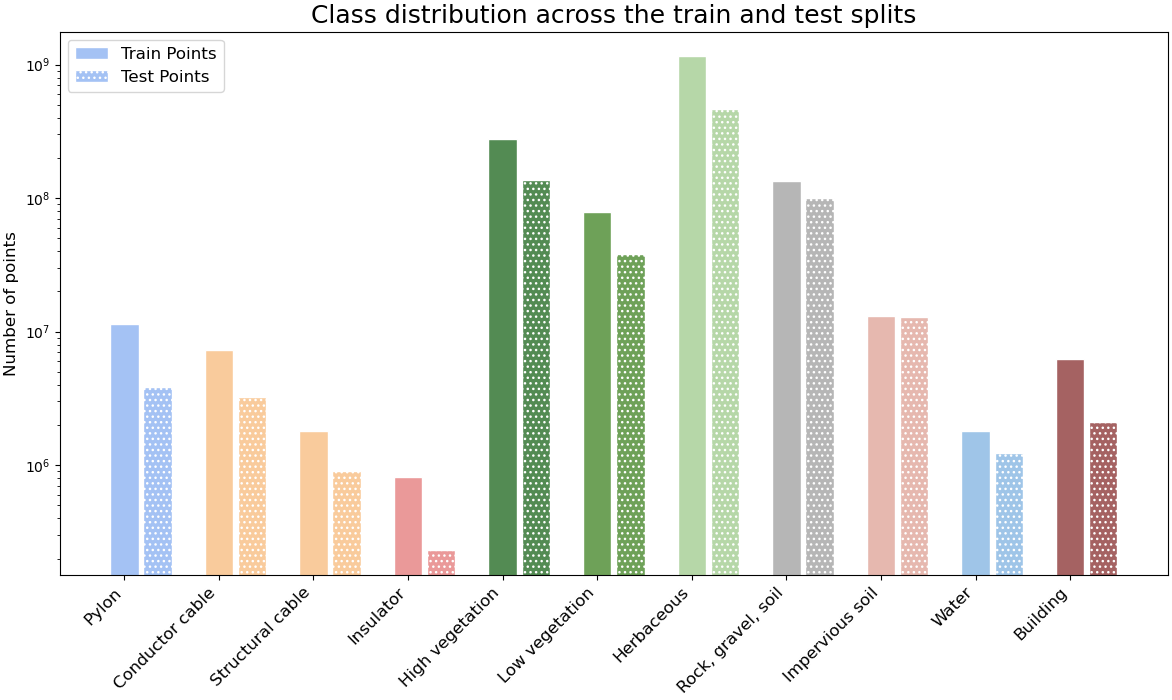}
\caption{Visual distribution of semantic class proportions across train and test sets. Each bar represents the number of 3D points of a class in each subset.}
\label{fig:split_visualization}
\end{center}

\subsection{Train/Validation Split}

To support model development and tuning, the training set was further split into a training and validation subset. Table~\ref{tab:split_val} summarizes class-wise statistics for this split, following the same structure as above.  This subdivision is proposed for reproducibility purposes and can be adjusted if needed.

\begin{center}
\small
\captionsetup{type=table}
\begin{tabular}{crrrccc}
\toprule
\textbf{Group ID} & \textbf{Train Points} & \textbf{Val Points} & \textbf{Total Points} & \textbf{\% Val/Total} & \textbf{Train Dist. (\%)} & \textbf{Val Dist. (\%)} \\ 
\midrule
0  & 8,643,791    & 2,846,313   & 11,490,104   & 24.8 & 0.7 & 0.7 \\
1  & 5,782,668    & 1,490,602   & 7,273,270    & 20.5 & 0.4 & 0.4 \\
2  & 1,370,331    & 441,091     & 1,811,422    & 24.4 & 0.1 & 0.1 \\
3  & 625,937      & 195,775     & 821,712      & 23.8 & 0.05 & 0.05 \\
4  & 160,763,512  & 117,764,269 & 278,527,781  & 42.3 & 12.4 & 29.7 \\
5  & 43,442,079   & 34,659,073  & 78,101,152   & 44.4 & 3.4 & 8.7 \\
6  & 968,689,542  & 186,527,777 & 1,155,217,319& 16.1 & 74.9 & 47.0 \\
7  & 87,621,550   & 47,404,508  & 135,026,058  & 35.1 & 6.8 & 11.9 \\
8  & 10,420,302   & 2,785,109   & 13,205,411   & 21.1 & 0.8 & 0.7 \\
9  & 310,240      & 1,496,976   & 1,807,216    & 82.8 & 0.02 & 0.4 \\
10 & 4,793,225    & 1,466,035   & 6,259,260    & 23.4 & 0.4 & 0.4 \\
\midrule
\textbf{TOTAL} & 1,292,463,177 & 397,077,528 & 1,689,540,705 & 23.5 & 100 & 100 \\
\bottomrule
\end{tabular}
\caption{Train/validation split statistics per semantic class (within the original training set).}
\label{tab:split_val}
\end{center}

\subsection{Zone Assignments}

Table~\ref{tab:split_zones} lists the specific zones assigned to each subset. While the train/test split is fixed, the subdivision of the training set into training and validation is a recommendation for reproducibility.

\begin{center}
\captionsetup{type=table}
\small
\begin{tabular}{cl}
\toprule
\textbf{Split} & \textbf{Zones} \\
\midrule
Train & t1z6a, t1z6b, t2z5, t3z3, t3z6, t3z7, t5a1, t5a3, t5a4, t5a5, t5b2, t5b3, t5b4, t5b6, t5c1, t5c2, t5c3, t6z2, t6z3, t6z4, t6z6 \\
Validation & t1z5b, t1z8, t3z4, t4z1, t5b1, t5b5 \\
Test & t1z4, t1z5a, t1z7, t3z1, t3z2, t3z5, t5a2, t6z1, t6z5 \\
\bottomrule
\end{tabular}
\caption{Assignment of data collection zones to the train, validation, and test sets.}
\label{tab:split_zones}
\end{center}

\section{Detail results of baselines}\label{app:baseline_results}

\subsection{Environment details}
All baselines were trained and evaluated on the same machine with the following configuration:
\begin{itemize}
    \item GPU: 4~x~NVIDIA A40 with 48 GB VRAM each
    \item CPU: 2~x~Intel Xeon Silver 4310 (48 cores)
    \item RAM: 512 GB
\end{itemize}
All methods were run using this machine with varying training configurations, 1~GPU only for ImageVote, SPT and Late Fusion, 4~GPU for PTv3 and DITR.

\subsection{Training and inference time}
The processing time for the training, validation, and test phases is reported in Table~\ref{tab:runtime_comparison} for each baseline, using the environment described above and the configurations provided on Hugging Face pages.

\begin{center}
\captionsetup{type=table}
\small
\begin{tabular}{lccccc}
\toprule
\textbf{Phase} & \textbf{ImageVote} & \textbf{SPT} & \textbf{Late Fusion MLP} & \textbf{PTv3} & \textbf{DITR} \\
\cmidrule{2-6}
Pre-processing & -&4~hr&24~min& 4~hr 30~min& 10~hr\\
Training       & 4~hr&20~hr&  38~min& 10~hr 10~min& 14~hr 50~min\\
Validation 3D  & 1~hr 40~min&42~min & 10~min& 5~min & 18~min\\
Test 3D        & 4~hr 20~min&1~hr 20~min&  22~min & 8~hr &  24~hr   \\
\bottomrule
\end{tabular}
\caption{Average runtime per phase for each baseline.}
\label{tab:runtime_comparison}
\end{center}

\newpage

\subsection{Per-class performance}
We report here the detailed per-class IoU scores on the test set for all baselines used in our study.
Table~\ref{tab:results_all_run} shows the average results over 3 training runs, including standard deviations for ImageVote, SPT, and Late Fusion.  
Table~\ref{tab:results_each_baseline_best_model} reports the performance of the best model (highest mIoU) selected for each method.

\begin{center}
\captionsetup{type=table}
\small
\begin{tabular}{lccc}
\toprule
\textbf{Class} & \textbf{ImageVote Baseline} & \textbf{SPT Baseline} & \textbf{Late Fusion MLP}\\
 &\textbf{IoU Test (\%) $\pm \sigma$} &  \textbf{IoU Test (\%) $\pm \sigma$} & \textbf{IoU Test (\%) $\pm \sigma$} \\
\cmidrule{2-4}
Pylon                     & 88.41 $\pm 2.35$& 90.80 $\pm 3.31$&  94.83 $\pm 0.04$   \\
Conductor cable           & 67.30 $\pm 1.76$& 90.29 $\pm 0.56$&  93.76 $\pm 0.46$   \\
Structural cable          & 42.21 $\pm 2.02$& 67.79 $\pm 1.92$&  81.73 $\pm 0.56$   \\
Insulator                 & 71.23 $\pm 0.13$& 78.60$\pm 2.03$&  84.72 $\pm 2.12$   \\
High vegetation           & 82.18 $\pm 1.25$& 85.95 $\pm 0.58$&  86.36 $\pm 2.39$   \\
Low vegetation            & 60.46 $\pm 2.11$& 54.88 $\pm 0.83$&  52.30 $\pm 5.01$   \\
Herbaceous vegetation     & 84.20 $\pm 0.19$& 84.42 $\pm 0.18$&  81.88 $\pm 0.80$   \\
Rock, gravel, soil        & 41.40 $\pm 1.97$& 38.34 $\pm 2.03$&  42.60 $\pm 0.48$   \\
Impervious soil (Road)    & 76.20 $\pm 3.19$& 71.85 $\pm 3.46$&  81.27 $\pm 0.79$   \\
Water                     & 66.72 $\pm 6.02$&  4.38 $\pm 0.64$& 52.62 $\pm 6.89$   \\
Building                  & 65.38 $\pm 2.24$& 57.86 $\pm 0.44$&  62.66 $\pm 1.10$   \\
\midrule
\textbf{Mean IoU (mIoU)}  & 67.79 $\pm 0.93$& 65.92 $\pm 0.69$& \textbf{74.07 $\pm 0.19$}  \\
\bottomrule
\end{tabular}
\caption{Mean per-class IoU over \textbf{3 runs}, comparison on test set for three baselines: ImageVote (image-only voting), SPT (SuperPoint Transformer), and Late Fusion MLP.}
\label{tab:results_all_run}
\end{center}

\begin{center}
    
\footnotesize
\captionsetup{type=table}
\begin{tabular}{lccccc}
\toprule
\textbf{Class} & \textbf{ImageVote Baseline} & \textbf{SPT Baseline} & \textbf{Late Fusion MLP} & \textbf{PTv3 w/ TTA and overlap} & \textbf{DITR w/ TTA and overlap}\\
 &\textbf{IoU Test (\%)} &  \textbf{IoU Test (\%)} & \textbf{IoU Test (\%)} & \textbf{IoU Test (\%)} & \textbf{IoU Test (\%)} \\
\cmidrule{2-6}
Pylon                     & 85.09 & 92.75 &  94.82 & 97.12 & 96.81  \\
Conductor cable           & 64.82 & 91.05 &  94.40 & 85.88 & 89.07  \\
Structural cable          & 45.06 & 70.51 &  82.52 & 53.22 & 57.80  \\
Insulator                 & 71.07 & 80.60 &  86.98 & 90.63 & 93.20  \\
High vegetation           & 83.86 & 85.15 &  83.08 & 88.30 & 88.81  \\
Low vegetation            & 63.43 & 55.91 &  47.64 & 33.93 & 41.99  \\
Herbaceous vegetation     & 84.45 & 84.64 &  80.75 & 91.72 & 90.05  \\
Rock, gravel, soil        & 38.62 & 40.63 &  42.89 & 51.88 & 44.26  \\
Impervious soil (Road)    & 80.69 & 73.57 &  80.26 & 76.63 & 79.49  \\
Water                     & 74.87 &  3.69 &  61.69 & 29.68 & 71.86  \\
Building                  & 68.09 & 57.38 &  61.40 & 60.49 & 70.26 \\
\midrule
\textbf{Mean IoU (mIoU)}  & 69.10 & 66.90 & 74.22 & 69.32 & \textbf{74.87} \\
\bottomrule
\end{tabular}
\caption{Per-class IoU, comparison on test set for \textbf{best model} from all baselines: ImageVote (image-only voting), SPT (SuperPoint Transformer), Late Fusion MLP, Point Transformer v3 (with overlap and TTA), DITR (with overlap and TTA).}
\label{tab:results_each_baseline_best_model}
\end{center}

\section{GridNet-HD Public Leaderboard}
To ensure reproducibility and standardized evaluation, we provide a leaderboard for the \emph{GridNet-HD} dataset, available at: \url{https://huggingface.co/spaces/heig-vd-geo/GridNet-HD-Leaderboard}.

The leaderboard evaluates 3D semantic segmentation model results on the test split, for which ground-truth annotations are not released. This ensures fair and blind evaluation across all submissions.

Submission Procedure:
\begin{itemize}
    \item Participants must generate predicted labels for all test point clouds.
    \item Export the classification field in .npz format (an example code to do that is provided directly on the leaderboard webpage), strictly maintaining the original point order provided by the dataset.
    \item Submit the 9 resulting .npz files on the leaderboard.
    \item Submitted results are evaluated using the mean Intersection over Union (mIoU) metric, which is computed server-side using the hidden test annotations.
\end{itemize}

Detailed submission instructions and format requirements are provided directly on the leaderboard page.

\bigskip

\end{strip}

\end{document}